\documentclass{article}
\usepackage{ijcai25}

\usepackage{times}
\usepackage{soul}
\usepackage{url}
\usepackage[hidelinks]{hyperref}
\usepackage[utf8]{inputenc}
\usepackage[small]{caption}
\usepackage{graphicx}
\usepackage{amsmath}
\usepackage{amsthm}
\usepackage{booktabs}
\usepackage{algorithm}
\usepackage{algorithmic}
\usepackage[switch]{lineno}

\usepackage{amssymb}
\usepackage{amsfonts}
\usepackage{multirow}
\usepackage{todonotes}
\usepackage{subcaption}
\usepackage[frozencache,cachedir=mintedcache]{minted}
\usepackage{forest}
\usepackage{marvosym}
\usepackage{tikz}
\usepackage{pgfplots}

\usepackage{cleveref}

\newcommand{\digit}[1]{\vcenter{\hbox{\includegraphics[height=10pt]{figures/#1}}}}

\newcommand{\imVar}[0]{\boxed{\texttt{?}}}

\newcommand{\newtext}[1]{\textcolor{black}{#1}}

\title{Declarative Design of Neural Predicates in Neuro-Symbolic Systems}

\author{
Tilman Hinnerichs$^1$
\and
Robin Manhaeve$^2$\and
Giuseppe Marra$^2$\And
Sebastijan Dumančić$^1$\\
\affiliations
$^1$TU Delft\hspace{1cm}
$^2$KU Leuven\\
\emails
\{t.r.hinnerichs, s.dumancic\}@tudelft.nl,\\
\{robin.manhaeve, giuseppe.marra\}@cs.kuleuven.be
}

\begin{document}

\maketitle

\begin{abstract}
    Neuro-symbolic systems (NeSy), which claim to combine the best of both learning and reasoning capabilities of artificial intelligence, are missing a core property of reasoning systems: Declarativeness. The lack of declarativeness is caused by the functional nature of neural predicates inherited from neural networks. We propose and implement a general framework for fully declarative neural predicates, which hence extends to fully declarative NeSy frameworks. We first show that the declarative
extension preserves the learning and reasoning capabilities while being able to answer arbitrary queries while only being trained on a single query type.
\end{abstract}

\section{Introduction}

The existing neuro-symbolic systems \cite{marra2024,hitzler2022}, which aim to unify the learning and reasoning capabilities of artificial intelligence (AI), lack one of the fundamental properties of reasoning systems -- their declarativeness.
Declarativeness is the property that makes reasoning systems universal question-answering engines.
That is, reasoning systems are procedures for computing answers to \textit{arbitrary queries} over \textit{arbitrary knowledge bases} where both are specified in a formal language.
For instance, logical reasoning systems, such as Prolog \cite{reason:SteSha86a} and Answer Set Programming \cite{lifschitz2019}, can answer arbitrary queries over a given knowledge graph (\textit{Who is a daughter of Lord Byron?, Who is the father of Ada Lovelace?}, and even aggregate queries such as \textit{ How many presidents did USA have in the past 100 years?}).
Similarly, probabilistic inference procedures can provide posteriors of arbitrary combinations of variables and evidence in a given Bayesian Network.

The capabilities of neuro-symbolic systems would significantly increase if they become declarative.
Most existing neuro-symbolic methods focus on problems that require reasoning about visual input; they use neural models to extract symbols from images and symbolic reasoning to reason about the content of the images.
If they become declarative, the same neuro-symbolic model could reason about the content of images but also inpaint an image while remaining logically consistent, create an image based on the logical description, provide a logical description of an image, and complete a partial logical description of an image, without a specialized model for each of these tasks.
Moreover, as the knowledge base is separated from the inference procedure, the knowledge base can be modified without compromising the capabilities of an AI agent.
That means that new symbolic knowledge about the images can be added without having to do any retraining.



The existing state-of-the-art systems, such as DeepProblog~\cite{manhaeve2018deepproblog} and SLASH~\cite{skryagin2022neural}, support declarativeness to a large extent.
However,  the declarativity breaks down with \textit{neural predicates}, the key abstraction of neuro-symbolic systems, which essentially evaluate the truthfulness of facts with neural networks.
For instance, a classic DeepProblog example uses neural predicates to map MNIST images to corresponding digits, $\texttt{digit(}\digit{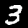}\texttt{,3)}$.
DeepProblog can answer any query that requires computation involving mapping a given image to a digit.
If a query requires mapping a given digit to an image, DeepProblog would not be able to answer such a query (while it would not struggle to answer such a query if the predicate in question is a logical predicate).
Recent DeepProblog extensions, VAEL \cite{misino2022vael}, SLASH~\cite{skryagin2022neural}, and DeepSeaProblog~\cite{desmet23}, show that such generative capabilities can be achieved, but 
without declarativeness.

\begin{figure}
    \centering
    \begin{subfigure}{0.48\textwidth}
    \begin{minted}{prolog}
addition(Im1,Im2,Z) :- digit(Im1,X2), 
        digit(Im2,Y2), Z is X2+Y2.
nn(mnist_net,[X],Y,[0,..,9])::digit(X,Y).
    \end{minted}   
    \caption{A DeepProblog program for addition}
    \end{subfigure}

    \begin{subfigure}{0.48\textwidth}
    \centering
    \begin{forest}
for tree={l sep=20pt, s sep=10pt}
[{\texttt{addition($\digit{mnist_3}$,$\boxed{?}$,7)}}
	[{\texttt{digit($\digit{mnist_3}$,X2)},\texttt{digit($\boxed{?}$,Y2)}, \texttt{7 is X2+Y2}}
		[{{\color{red}\texttt{digit($\boxed{?}$,4)}}, {\color{green}\texttt{digit($\digit{mnist_3}$,X2)}}},edge label={node[align=left,midway,left=0.55cm] {$ X2=3$\\ $Y2=4 $}} 
			[$\bot$,edge label={node[align=left,midway,left] {$ \boxed{?}=$\Lightning }}]
		] 
		[{$\dots$}]
		[$\dots$,edge label={node[align=left,midway,right=0.5cm] {$X2=\dots$\\ $Y2=\dots $}} ] 
	]
]
\end{forest}
    \caption{Resolution tree for the query \texttt{addition($\digit{mnist_3}$,$\boxed{?}$,7)}}
    \end{subfigure}
    \caption{\textbf{Unification in DeepProblog breaks when tasked to generate an image.} DeepProblog iteratively tries to match either the current facts with rules from the program in (a) or variables with values. 
    After \texttt{digit($\boxed{?}$,4)} is generated it cannot assign a value to $\boxed{?}$, breaking the resolution algorithm.
    }
    \label{fig:digit_resolution_tree}
\end{figure}

The lack of declarativeness is caused by the functional nature of neural networks integrated into reasoning frameworks. 
First, neural networks are functions, not relations, with clearly defined inputs and outputs that cannot be exchanged.
Reasoning systems are instead based on relations, where arguments can take arbitrary roles of inputs and outputs.
Second, neural networks are discriminators, meaning that they learn a conditional distribution of labels given features but do not model properties of the feature space.
\newtext{Even the generative architectures, such as variational auto-encoder \cite{VAE2019Kingma} and diffusion models \cite{DiffusionYangZSHXZZCY24}, operate as functions, generating, e.g., images from noise or latent representations, despite not performing a classification task.}
For instance, a perfect MNIST digit classified does not model a particular digit's distribution of valid images. 

\newtext{The main consequence of this is that neural predicates
make unification, a key operation in logical reasoning systems that matches possible values to variables, impossible.
Such reasoners iteratively try to match current facts with possible rules to derive their truth. 
Neural predicates can only be matched in the direction they were trained on. 
For example, in the vanilla DeepProblog, we can only map images to their label.
Hence, the resolution fails when generating images instead of classifying them, as shown in \cref{fig:digit_resolution_tree}.
}

In this paper, we \newtext{introduce} a general framework for designing declarative neural predicates.
\newtext{In essence, our two contributions focus on enabling unification to operate over infinite domains that are difficult to characterize precisely, such as sensory input in the form of images or sounds. }
First, the central idea behind our approach is to design neural predicates around \textit{prototypes} and prototype networks \cite{xu2020attribute}.
Designing neural predicates around prototypes forces them to learn concepts instead of mapping from sensory input to labels, avoiding the issues with unification.
Second, we demonstrate how a relational interpretation of the encoding-decoding scheme, relating instances to their prototypes, resolves the relations-vs-functions dichotomy.
Because prototypes effectively learn a domain, any argument of the corresponding predicate can be \textit{sampled} from the prototype and thus unified with a variable.
We design the declarative neural predicates in such a way that our proposed framework does not require new special machinery, but it reuses the inference procedures of DeepProblog.

We implemented our framework within DeepProblog and evaluated it on the subset of DeepProblog \cite{manhaeve2018deepproblog} problems for which prototype interpretation makes sense.
We demonstrate that our declarative neural predicates learn prototypical instances from data, achieve comparable performance to DeepProblog despite solving a more complex task, and expand DeepProblog's capabilities by answering arbitrary queries involving neural predicates.

\section{Background}

\subsection{Declarative problem solving}

We start by recapping the idea of declarative problem-solving for an uninitiated reader, illustrating the types of behaviour we want to recover for neuro-symbolic systems.
We focus on logic programs~\cite{reason:SteSha86a} as they are the basis for this work and DeepProblog.

Logic programs reason over knowledge bases, which state things we know.
The knowledge bases contain facts, the logical statements we know are true.
For example, they could state the relative sizes of certain animals.

\begin{minted}{prolog}
bigger(elephant, horse).
bigger(horse, donkey).
bigger(donkey, dog). 
bigger(donkey, monkey).
\end{minted}

Knowledge bases also contain rules we can use to deduce further facts. 
For example, the following rule allows us to combine the previously stated facts towards their relative close.

\begin{minted}{prolog}
is_bigger(X, Y) :- bigger(X, Y).
is_bigger(X, Y) :- bigger(X, Z), 
                    is_bigger(Z, Y).
\end{minted}

Logic programs can answer different queries with these facts and rules in the knowledge base.
We can query the entire knowledge base for the facts directly.
For instance, whether an elephant is bigger than a horse, which would correspond to the following query \texttt{?-is\_bigger(elephant,horse).}\footnote{the character \texttt{?-} is a Prolog convention indicating that we are interested in establishing the truth value of a logical statement following the character}
We could also ask which animal is bigger than a dog, \texttt{?-is\_bigger(X,dog).}\footnote{capitalized symbols like \texttt{X} stand for variable, while lowercase symbols stand for entities like \texttt{dog} stand for entities in the knowledge base}, or which animal is smaller than a donkey, \texttt{?-is\_bigger(donkey,X)}. 
We could even ask for all pairs of animals such that the first one is bigger than the second one, \texttt{?-is\_bigger(X,Y).}.
These kinds of queries can be posed about any predicate
in the knowledge base, not prespecified ones.
The ability to do all of these is referred to as declarative behaviour.

An essential consequence of declarativeness is that the knowledge base can be modified freely.
For instance, if we add the fact \texttt{bigger(whale,elephant)}, a logic programming engine would immediately produce correct answers.
Similarly, if we add a new predicate, for instance, \texttt{swims(whale)}, the engine would immediately be able to answer queries involving the \texttt{swims/1} predicate.

We want to unlock this kind of behaviour in neuro-symbolic systems.

\subsection{DeepProblog}

We briefly summarise the basics of DeepProblog, which we use as a basis for our contributions.

DeepProblog extends Problog \cite{de2007problog,kimmig2011implementation}, the probabilistic extension of Prolog, enriching facts with probabilities of being true.
A Problog program consists of (1) a set of ground probabilistic facts $\mathcal{F}$ written as $p :: f$ for fact $f$ with probability $p$ of being true,  and a set of clauses $\mathcal{R}$. 


Probabilistic facts and rules jointly (with their entailments) define possible worlds.
The probability of a possible \textit{world} $w_F$ is described by a subset of facts $F\subseteq \mathcal{F}$ and is defined as $w_F = F \cup \{f_\theta|\mathcal{R} \cup F \models f_\theta \text{ and } f_\theta \text{ is ground}\}$, where each fact $f\in F$ corresponds to an independent Boolean random variable with probability $p_f$. The probability of a world $P(w_F)$ is defined by 

\begin{equation}
P(w_F) = \prod_{f\in F} p_f \prod_{f\in \mathcal{F}\backslash F}(1-p_f) 
\end{equation}

The probability of a query $q$ is then defined by the sum of probabilities where $q$ holds, i.e. 

\begin{equation}
    P(q) = \sum_{F \subseteq \mathcal{F}:q\in w_F} P (w_F)
\end{equation}


DeepProblog \cite{manhaeve2018deepproblog} extendProblog with neural predicates $\{q\}$ which determine the probability of a probabilistic fact \textit{dynamically}.
This makes it possible to conveniently model probabilistic facts defined over technically infinite spaces, like images or sounds.
Neural predicates are modeled using ground annotated disjunctions (nADs) of the form
\begin{equation}
nn(m_q , \vec{t}, \vec{u}) :: q(\vec{t}, u_1); ...; q(\vec{t}, u_n) :- b_1, ..., b_m
\end{equation}
where $b_i$ are the atoms, $\vec{t}=t_1,\dots,t_k$ is a vector of inputs of the neural network, $u_1$ to $u_n$ are the network's possible output values and $m_q$ is the identifier of the used network.
This means that neural predicates effectively model a conditional distribution of labels given an image.

\subsection{Prototype networks}
Prototypes make the representation of each category or class of data explicit.
Prototype networks \cite{SnellSZPrototypeNetworks2017} represent each possible concept using a prototype in a latent space.
To classify, they learn an embedding function that maps inputs to the same latent space.
An input is assigned to a class by applying the embedding function and computing the distance to the class' prototype.

\section{Related work}

\paragraph{Neuro-symbolic systems.} 
Many neuro-symbolic systems have been introduced over the years: DeepProblog \cite{manhaeve2018deepproblog}, Scallop \cite{huang2021scallop}, NeurASP \cite{yang2023neurasp}, Logic Tensor Networks \cite{badreddine2022logic}, and NeuPSL \cite{pryor2023deep} and many others.
These systems predominantly use neural predicates to process sensory data, such as images, into a symbolic interpretation suitable for reasoning.
None of these systems are fully declarative.
However, our contributions can be directly incorporated into any of these systems. 

\paragraph{Generative capabilities in neuro-symbolic systems.}
Several recent systems are directly related to our work.
SLASH \cite{skryagin2022neural} and VAEL \cite{misino2022vael} demonstrate the ability to answer queries requiring image generation. 
However, their design remains inherently functional.
They must be explicitly trained to solve the particular generative task and cannot answer arbitrary queries without retraining. 
Vieira \cite{Li2024} is a system for programming with (generative) language models that focuses on combining them, but any declarative properties are prescribed, not learned from data.
\nocite{jiang2020generative}

DeepSeaProblog~\cite{desmet23} is the work most closely related to ours.
DeepSeaProblog extends DeepProblog with support for continuous distributions.
Though DeepSeaProblog does not tackle the issue of declarative neuro-symbolic systems, it provides a suitable language for modeling declarative neural predicates.
One of the demonstrations presented in the paper uses an auto-encoding structure similar to our proposal. 
However, DeepSeaProblog does not establish a basis for turning any neural predicate into a declarative one.


%
%
%

\section{Methodology}

We now explain our framework for declarative neural predicates.
\newtext{We will use the \texttt{digit/2} predicate from the MNIST example as a running example, where digits are matched with their images.}

Our framework is based on two key ideas.
First, to answer arbitrary queries that involve neural predicates, there exists a set of \textit{canonical queries} to which other queries can be reduced.
In our example, the canonical set of queries that we want to be able to answer without (re)training is
\begin{itemize}
    \item $\texttt{digit(}\digit{mnist_3}\texttt{,3)}$: is this an image of the digit 3?;
    \item $\texttt{digit(}\digit{mnist_3}\texttt{,?)}$: what is the digit in this image?;
    \item $\texttt{digit(}\imVar\texttt{,3)}$: what is an image associated with the digit 3?; and 
    \item $\texttt{digit(}\imVar\texttt{,?)}$: what are valid groundings of this predicate?
\end{itemize}
More complex queries involving more neural predicates and combinations thereof would reduce to combining this set of queries.
Our framework contributes inference procedures to answer all these queries.

Second, the main challenge in answering canonical queries is that the non-symbolic arguments of neural predicates do not come with \newtext{concretely} defined domains.
However, if a user poses a query involving a domain of images, the reasoning engine needs to know which values can appear in that place.
We overcome this problem by structuring predicates around prototypes, which learn the domains of non-symbolic arguments from data.

We present a framework rather than a fully automatable process; the user must still decide how to structure the neural predicates around prototypes.
However, the construction can be done automatically under some fairly general assumptions.
Our framework is designed to avoid changes to the internal mechanisms in (Deep)Problog.
The proposed framework changes the semantics of DeepProblog programs but preserves the inference procedure.

We first explain the design principles behind declarative neural predicates and \newtext{operations over them}.
\newtext{The formal definition of their semantics is in appendix \ref{sec:app:Semantics}.}



\subsection{Neural predicates with prototypes}

\begin{figure*}[th]
    \centering
    \input{figures/encode_decode}
    \caption{\textbf{Procedures to answer the four canonical queries.} Each prototype is shown in latent space and consists of a mean $p_i$ and a distribution illustrated in a darker shade. 
    The arrows represent different computations, as shown in the legend. 
    (a) Both image and prototype are mapped to the latent space. The probability is their distance in latent space.
    (b) The image is encoded, compared to all prototypes, and the closest is used to answer the query.
    (c) The prototype is mapped to the latent space, an instance is sampled and decoded back to image space.
    (d) All possible groundings are generated. The prototype is fetched for each possible grounding, and an instance gets sampled and decoded back to image space.
    }
    \label{fig:digit_encode_decode}
\end{figure*}

Instead of treating neural predicates as mappings from images to symbols, we \newtext{task neural predicates with capturing \textit{prototypes} and relating images to them.} 
The goal of a prototype is to learn \textit{true groundings} of a neural predicate.
All images of digit 1 should be associated with the prototype, but no other images.
We impose five modelling assumptions on neural predicates.

\paragraph{Assumption 1.}
We assume a fixed number of prototypes.
In our example, we assume ten prototypes corresponding to one of the digits each.
We assume that the user gives the number of prototypes and leave the problem of automatically finding them for future work.

\newtext{
While we assume a fixed number of prototypes, the prototypes themselves are learned from data with prototype networks \cite{xu2020attribute}.
Prototypes are usually captured in \textit{latent space}, not the same representation spaces as the original inputs; we follow the same principle.
}

\paragraph{Assumption 2.}
The prototypes are responsible for grounding \textit{non-symbolic arguments} of the predicate, that is, the arguments that are not expressed as logical symbols, such as images, sound, or any other sensory input.
Considering our MNIST example, as each prototype is associated with one of the ten digits, the role of the prototypes is to produce groundings of images related to it, i.e., the images of a particular digit. 

\paragraph{Assumption 3.}
One neural predicate models exactly one non-symbolic argument.
A predicate involving two or more non-symbolic arguments can be modeled through a clause combining several neural predicates.
For instance, assume we want to have a predicate \texttt{visual\_addition/3} which takes three arguments, all being images, such that the third image is the sum of the digits in the first images. This assumption means that we will not model such a predicate as a single neural predicate operating on one or three prototypes but rather as a composition of three neural predicates focusing on individual images:

\begin{minted}{prolog}
visual_addition(Im1, Im2, Im3):- 
        digit(Im1,D1), digit(Im2,D2)
        digit(Im3,D3), D3 is D1 + D2
\end{minted}


\paragraph{Assumption 4.}
Symbolic arguments of neural predicates are associated with prototypes and can be determined from them. 
In our MNIST example, each prototype has a digit associated with it; therefore, by knowing the prototype an instance belongs to, we can determine the second argument of the \texttt{digit/2} predicate, i.e., the corresponding digit.
$n$-ary predicates can be handled similarly.

\newtext{
We do not assume that there has to be a 1-to-1 mapping between a prototype and symbolic arguments of a predicate.
One prototype can map to more than one combination of symbolic arguments.
}


\paragraph{Assumption 5.}
We assume that prototypes are mutually exclusive, i.e., every non-symbolic instance can belong to only one prototype.
In DeepProblog terms, the prototype membership is an annotated disjunction. 
This assumption is not restrictive: the original formulation of neural predicates also assumes that neural predicates declare annotated disjunction, meaning that one image can have only one label attached to it.  
In our case, an MNIST image can only belong to one prototype, and it is up to the DeepProblog engine to decide which one.

We model prototype membership as an annotated disjunction where the probability of assignment to a particular prototype is proportional to the distance from the learned prototype (\cref{fig:declneurpred}, lines 1--5).
The prototypes form a mixture model, and the annotated disjunction ensures that a particular non-symbolic instance belongs to one component of the mixture (one prototype) at a time.
This formulation closely follows the standard Gaussian mixture model~\cite{Reynolds2009}.

\begin{figure*}[th]
    \centering
    \begin{subfigure}{\textwidth}
    \begin{minted}[linenos, xleftmargin=20pt, highlightlines={15-16, 19-20}, highlightcolor=red!15]{prolog}
P1::digit(Image,1); P2::digit(Image,2); P3::digit(Image,3) :- 
        prototype(1, Prot1), prototype(2, Prot2), prototype(3, Prot3), 
        prototype_match(Image, Prot1, P1),
        prototype_match(Image, Prot2, P2),
        prototype_match(Image, Prot3, P3).

prototype_match(Image, Prot, P) :- encode(Image, Prot, P1), 
        decode(Prot, Image, P2), mul(P1, P2, P). 

encode(Image, Prot, P) :- ground(Image), nn_encoder(Image,Latent2), 
        lat_similar(Prot, Latent2, P). 
encode(Image, Prot, P) :- var(Image), sample(Prot, Sample), 
    nn_decoder(Sample, Image), lat_similar(Prot, Sample, P).

decode(Prot, Image, P) :- ground(Prot), sample(Prot, Sample), 
        nn_decoder(Sample, Image2),  im_similar(Image, Image2, P).
decode(Prot, Image, 1.0) :- var(Prot), nn_encoder(Image, Latent).

im_similar(X,X, 1.0).
im_similar(Image1, Image2, P) :- Image1 \= Image2, 1-mse(Image1, Image2, P).
lat_similar(X,X, 1.0).
lat_similar(Lat1, Lat2, P) :- Lat1 \= Lat2, likelihood(Lat1, Lat2, P).
    \end{minted}
    \caption{Training program}
    \end{subfigure}

\begin{subfigure}{\textwidth}
    \begin{minted}[linenos, xleftmargin=20pt, highlightlines={1}, highlightcolor=green!15,]{prolog}
decode(Prot, Image, 1.0) :- ground(Prot).
    \end{minted}
    \caption{Substitution used during inference.}
\end{subfigure}
    
    \caption{\textbf{Declarative neural predicate for the MNIST digit example with three digits only.} (a) The first clause is an annotated disjunction modeling prototype membership of an image \texttt{I}. It relates it to each prototype through \texttt{prototype\_match/3}. The predicate \texttt{prototype/2} relates a particular prototype to its latent representation. 
    The \texttt{prototype\_match/3} is essentially a probability fact capturing the likelihood that an image belongs to the prototype. 
    This is achieved by comparing the latent projection of an image to the prototype space and the prototypes themselves. 
    The prototypes are modeled as Gaussian distributions; the similarity to a prototype is computed as the likelihood of sampling a latent representation of an image from the distribution defined by the prototype. 
    The predicates \texttt{im\_similar/2} and \texttt{lat\_similar/2} compute similarities in the image space (as 1 - mean squared error between the images) and the probability of a latent vector being sampled from the Gaussian distribution defined by the prototype, respectively.
    \newtext{(b) During inference, the snippet marked in red is substituted by the green code snippet. 
    As the decoder is trained already, we can shorten the \texttt{decode} rule and remove \texttt{im\_similar}.
    }
    }
    \label{fig:declneurpred}
\end{figure*}

\subsection{Relational operations on prototypes}

By design, prototypes are learned from data and, therefore, realize the domain of valid instances. 
An important consequence is that we can now \textit{sample} instances from a prototype.
By doing so, we overcome the previously outlined problem of unification. If we have to unify a variable standing in for a non-symbolic argument, we can sample from the corresponding prototype.

This ability to sample instances from a prototype allows us to answer the four canonical queries \newtext{(see \cref{fig:digit_encode_decode})}:
\begin{itemize}
    \item  $\texttt{digit(}\digit{mnist_3}\texttt{,3)}$: to answer this query, we map the given image to the latent space and compute the distance of the mapping to the prototype associated with digit 3;
    \item $\texttt{digit(}\digit{mnist_3}\texttt{,?)}$: to answer this query, we map the image to the latent space, find the closest prototype, and determine the corresponding digit;
    \item $\texttt{digit(}\imVar\texttt{,3)}$: to answer this query, we determine the prototype associated with the digit 3, sample an instance for the prototype, and unify it with the image variable, 
    \item $\texttt{digit(}\imVar\texttt{,?)}$: to answer this query, we iterate over all prototypes and sample an instance from them. The instance is then decoded into an image.
\end{itemize}



The inference procedures for the canonical set of queries demonstrate that we must move flexibly between non-symbolic instances and the prototypes.
To do so, we introduce a relation \texttt{prototype\_match/3} which relates the image, a prototype, and the probability that the image is associated with the prototype.
This can be seen as a relation interpretation of the encoding-decoding scheme standard in (variational) autoencoders. 

The relation defined in line 7 of \cref{fig:declneurpred} first states the obvious: the given images embed into the prototype, and the prototype decodes the image.
The interesting part is in these two inner relations.
The relation \texttt{encode(Im,Prot,P)} states that an image \texttt{Im} belongs to a prototype \texttt{Prot} with probability \texttt{P}:

\begin{minted}{prolog}
encode(Image, Prot, P):-
    ground(Image),nn_encoder(Image, Lat),
    lat_similar(Prot,Lat, P).
encode(Image, Prot, P) :- var(Image), 
    sample(Prot, Sample), 
    nn_decoder(Sample, Image),
    lat_similar(Prot, Sample, P).  
\end{minted}

If an image is given, it is mapped to the prototype space through the prototype network \texttt{nn\_encoder/2}. 
The probability is determined by the similarity of the encoding to the prototype.
If the image is not given, we sample from the given prototype to generate an image.

The \texttt{decode/3} relation relates prototypes to images from their domain and is defined similarly to the \texttt{encode/3} relation:


\begin{minted}{prolog}
decode(Prot, Image, P) :- ground(Prot), 
    sample(Prot, Sample), 
    nn_decoder(Sample, Image2),  
    im_similar(Image, Image2, P).
decode(Prot,Image,1.0):-
    var(Prot), nn_encoder(Prot,Image).    
\end{minted}

\paragraph{Training objective.}
The \texttt{im\_similar/2} predicate measures the similarity between the given image and the one generated from a prototype.
This predicate essentially defines a \textit{reconstruction loss}, pushing the decoder to learn to generate images similar to the ones in the data.
This predicate is unnecessary during inference, and we remove it from the program. \footnote{It is possible to include reconstruction loss outside of the DeepProblog program; this is a simple and easy way to achieve this. }



\paragraph{Prototype membership probabilities.}
The missing part of the framework is the probabilities of probabilistic facts that associate instances with prototypes.
To do so, we design prototypes as multivariate Gaussian distributions parametrised by a mean and standard derivation, following the design of variational auto-encoders~\cite{Kingma2014}.
This design allows a prototype to capture the variance present in images that should be associated with it.
The probability of an instance belonging to a prototype is then defined as a \textit{normalised likelihood} of an instance being sampled from a Gaussian distribution captured by the prototype.


\paragraph{Implementing encoder and decoder.}
The framework presented so far does not depend on the exact structure of encoders and decoders.
While in this work, we rely on a variational auto-encoder, any generative network can be used as a decoder.
For instance, to use diffusion models \cite{DiffusionYangZSHXZZCY24} or generative adversarial networks \cite{GANS2014Goodfellow}, the prototypes can be interpreted as capturing a \textit{distribution of input noise} to be fed into these models.


\paragraph{Unification over infinite spaces.}
An attentive reader might have noticed the prototypes described so far do not reduce the domain.
They learn what typical instances look like, but the domain remains continuous and infinite.
This remains a problem for unification because if it encounters a variable corresponding to an image, it would try to fill it with infinite possibilities.
The way we have structured declarative neural predicates helps to avoid this issue: a possible image is sampled from a prototype, and we limit our framework to retrieving a single sample.

\section{Experimental evaluation}

The experimental evaluation of our framework focuses on verifying that prototype-based neural predicates can be effectively learned from data and that they extend the capabilities of neuro-symbolic systems with declarative reasoning.
We separate the evaluation into the following research questions:
\begin{itemize}
    \item[] \textbf{RQ1}: Can prototype-based neural predicates be learned from direct supervision?
    \item[] \textbf{RQ2}: Can prototype-based neural predicates be learned from distance supervision?
    \item[] \textbf{RQ3}: Can prototype-based DeepProblog answer arbitrary queries?
\end{itemize}

\subsection{Methodology}

\paragraph{Tasks.}
To evaluate our framework, we use a classical problem of MNIST addition introduced in  DeepProblog~\cite{manhaeve2018deepproblog} and DeepStochlog~\cite{winters2022}.
These tasks allow for a meaningful interpretation of prototypes, which capture the digits images represent.
More specifically, we use the following tasks:
\begin{itemize}
    \item \textbf{Direct supervision} (\texttt{digit/2}):  we learn prototype-based neural predicates from examples such as $\texttt{digit(}\digit{mnist_3}\texttt{,3)}$. This is essentially a sanity check to assess the performance of our new neural predicates, which have more complex tasks to solve than the original DeepProblog.
    \item \textbf{Distant supervision} (\texttt{add/3}): we learn neural predicates from examples such as $\texttt{addition(}\digit{mnist_3}\texttt{,}\digit{mnist_3}\texttt{,6)}$, matching the DeepProblog setup. 
    \item \textbf{Unseen queries - simple} (\texttt{digit/2} and \texttt{add/3}): we use the model from RQ1 and query it with queries it has not been trained to answer. More precisely, we use queries $\texttt{?-digit(}\imVar\texttt{,?).}$ (return images for all possible digits), and $\texttt{?-addition(}\imVar\texttt{,}\imVar\texttt{,7)}$ (return images of all combinations of digits that sum to 7). No other system can answer these queries without being explicitly trained to do so.
    \item \textbf{Unseen queries - complex} (\texttt{multi\_add/3}): starting from the model trained in RQ1, we prompt it to solve $4$-digit addition. We randomly substitute $4$ images with variables to sample $n=100$ declarative queries (see Appendix \ref{sec:app:experiments} for an example). Again, no other system is able to solve this task.
\end{itemize}

We train our declarative and original DeepProblog version on 60000 examples for\texttt{digit/2} and 30000 examples for \texttt{add/3}  and test on a separate test set of 10000 samples. 

\paragraph{Performance metrics.}
For the first two tasks, we use classification accuracy to measure success.
For the two tasks involving unseen queries, we calculate the accuracy of answers in the following way. 
For every image our system returns as an answer, we find the most similar image from the training set (using the mean squared error as a distance) and retrieve the corresponding label. 
We then use these labels to calculate the accuracy of the answers, denoted as \textit{generative accuracy}.
We opt for this measure rather than the choice of prototypes, as the system's performance is judged based on the final output it generates, not the want it intends to generate.


\paragraph{Implementation details.}
Our implementation uses a neural architecture similar to DeepProblog, except that the last layer of the encoder maps to a prototype rather than a label. 
We use a fully connected network as a decoder.
\newtext{Similar to DeepProblog, we use binary cross-entropy as the loss function.}
One of our framework's hyperparameters is the similarity measure choice, comparing instances to prototypes. 
For our variational autoencoder (VAE) setup, we use likelihood to describe similarity in latent space, i.e., between prediction and respective prototypes.
To measure similarity in image space, we use the inverted mean squared error (mse), that is $\texttt{1-mse(Im1,Im2)}$.
Digit classification and addition tasks were repeated 5 and 3 times, respectively, and average measures were reported.

\subsection{Results}

\subsubsection{RQ1: Learning prototypes from direct supervision}
\label{sec:exp:learning}
\newtext{
When learned from direct supervision, prototype-based neural predicates are as effective as their functional counterpart (\cref{tab:ex_accuracies}).
This result demonstrates that declarative neural predicates do not necessarily sacrifice performance and can be effectively learned, despite solving a more challenging task by capturing domains through prototypes.
}

\subsubsection{RQ2: Learning prototypes from distant supervision}
Similarly to the previous experiment, our declarative neural predicates perform comparatively to their functional counterpart when trained on distant supervision.
This result is especially interesting because distant supervision arguably provides a less clear signal to learn the domains of instances. 




\begin{table}[t]
\centering
\caption{Declarative neural predicates achieve comparable performance when trained on direct supervision and perform well but noticeably worse when trained on distant supervision.}
\begin{tabular}{@{}lrr@{}} 
\toprule 
\textbf{Model} & \multicolumn{2}{c}{\textbf{Accuracy}} \\
\cmidrule{2-3}
 & \texttt{digit/2} & \texttt{add/3} \\ 
\midrule
DeepProblog    & 98.7\% & 95.6\%\\
DeclDeepProblog & 98.4\% & 94.2\%\\ 
\bottomrule
\end{tabular}
\label{tab:ex_accuracies}
\end{table}

\subsubsection{RQ3: Declarative queries}
The most interesting aspect of our evaluation is the unseen queries.
The results (\cref{tab:decl_accuracies}) demonstrate that our declarative DeepProblog can successfully answer the simple queries, \texttt{digit/2} and \texttt{add/3}, obtaining high generative accuracy. 
Interestingly, while the generative accuracy is imperfect, our system always picks the correct prototype.
The generated images, sampled from the selected prototype, look more similar to those generated from another prototype.
This issue might be overcome by using a more powerful generative model or training longer.

\begin{figure}[t]
    \centering
    \includegraphics[width=0.08\textwidth]{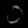}
    \includegraphics[width=0.08\textwidth]{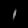}
    \includegraphics[width=0.08\textwidth]{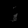}
    \includegraphics[width=0.08\textwidth]{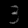}
    \includegraphics[width=0.08\textwidth]{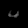}
    \includegraphics[width=0.08\textwidth]{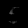}
    \includegraphics[width=0.08\textwidth]{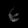}
    \includegraphics[width=0.08\textwidth]{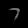}
    \includegraphics[width=0.08\textwidth]{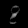}
    \includegraphics[width=0.08\textwidth]{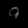}
    \caption{Images learned by prototypes, sorted from 0 to 9. They resemble key distinguishing features of digits but not their full appearance.}
    \label{fig:digit_grounding}
\end{figure}

\cref{fig:digit_grounding} shows the images sampled from each prototype when querying $\texttt{?- digit(}\imVar\texttt{,?).}$.
The prototypes pick up on the distinguishing features of digits rather than their whole appearance.
This could result from using mean squared error as the reconstruction loss or not having a sufficiently powerful sampling architecture.

\newtext{Our system also performs well on the multi-digit addition problem \cref{tab:decl_accuracies}, having the generative accuracy well above random guessing.
The actual performance is much lower than in the previous cases, but this is to be expected as the error accumulates with more digits. 
However, similar to the previous case, most errors are caused not by selecting a wrong prototype but by generating an image that looks more similar to another digit.
Again, we note that no other system is able to answer all of these queries simultaneously.
}




\begin{table}[t]
\centering
\caption{Declarative DeepProblog accurately generates images for queries it has not been trained on.}
\begin{tabular}{@{}lrrr@{}} 
\toprule 
\textbf{Model} & \multicolumn{3}{c}{\textbf{Accuracy}} \\
\cmidrule{2-4}
& \texttt{digit/2} & \texttt{add/3} &\texttt{multi\_add/9} \\ 
\midrule
DeclDeepProblog & 88.1\% & 81.5\%& 62.2\%\\ 
\bottomrule
\end{tabular}
\label{tab:decl_accuracies}
\end{table}


\section{Conclusion}

This work tackles an open problem within neuro-symbolic systems: How can we make them fully declarative?
Focusing on the DeepProblog family of neuro-symbolic systems, we outline challenges that prevent these systems from being fully declarative, all arising from the functional nature of neural networks.
To overcome these limitations, we introduce a framework for making any neural predicate, the key abstraction in DeepProblog, fully declarative.
Our framework introduces a specific design of neural predicates around prototypes; this design allows them to regain declarativeness. 
Our initial experiments demonstrate the potential of the proposed framework: our neural predicates demonstrate similar performance to a functional one while being able to answer arbitrary queries.

Many open challenges remain if we are to reach the goal of fully declarative neuro-symbolic systems.
One of the biggest challenges is the scalability.
We have noticed that incorporating exclusive membership to prototypes ensures fast convergence; however, each training iteration takes twice as much time.
Furthermore, distant supervision makes learning more challenging.
We suppose this is the case because distant supervision does not provide a 1-to-1 assignment between images and prototypes; the example $\texttt{addition(}\imVar\texttt{,}\imVar\texttt{,5)}$ does not give a unique assignment to the first two arguments but several options are possible: \texttt{0+5=5}, \texttt{5+0=5}, \texttt{1+4=5}, \texttt{4+1=5}, \texttt{2+3=5}, \texttt{3+2=5}.
Finally, we have assumed a fixed number of prototypes; automatically learning the number of necessary prototypes would make the entire framework more usable.

\bibliographystyle{named}
\bibliography{ijcai25}

\appendix

\clearpage
\appendix

\section{Semantics}
\label{sec:app:Semantics}

We show the semantics of the neural predicate in the MNIST case, but the explanation holds beyond it.
The neural predicate $p(X,d)$ is a joint probability distribution:
\[p(X,d) = \int p(X | z) p(z | d) p(d) dz\]

where $X$ is an image, $z$ is a latent embedding and $d$ is a digit.
Here:
\begin{itemize}
    \item $p(d)$ is a prior over digits; we model it as a uniform categorical distribution over the digits; 
    \item $p(z|d)$ is a multivariate gaussian distribution
    ; it is parameterized by a simple table mapping $d$ to the mean and standard deviation of the Gaussian; this is the prototype associated with $d$.
    \item $p(X|d)$ is a set of independent Gaussian distributions of pixel intensities parameterized by a deep neural decoder.  
\end{itemize}

Given the model, we will now show how to perform different inference tasks on this model.

\paragraph{Generative query.}The purely generative query $\texttt{digit(}\imVar\texttt{,?)}$ can be  answered by sampling from the distribution $p(X,d)$ following the conditional factorization.

\paragraph{Conditional query on digits.}
The conditional query $\texttt{digit(}\imVar\texttt{,3)}$ can be answered as the previous one but starting from the provided digit.

\paragraph{Conditional query on image.}
To answer such a query, we need to introduce two approximations to the intractable posterior distributions $p(z | X)$ and $p(d | z)$. We introduce a variational Gaussian approximation $q(z|X)$, parameterized by a neural network encoder, and a categorical digit distribution: 

\[q(d|z) = \frac{1}{Z} e^{-\frac{d(z ,z_d)}{T}}\]

where $d$ is a distance function in embedding space and $Z = \sum_\delta e^{-\frac{d(z ,z_\delta)}{T}}$ is a normalization factor. Therefore,

\[p(d|X) \approx \int q(d|z)q(z|X) dz\]

The corresponding query can thus be answered by sampling from $q(z|X$) and $q(d|z)$. Notice that $q(d|z)$ shares the same parameters  (prototypes) $z_d$ with $p(z|d)$, thus making the most probable digit $d$ the one whose prototype $z_d$ in $p(z,d)$ is the closest.

\paragraph{Likelihood query.}
The final query $\texttt{digit(}\digit{mnist_3}\texttt{,3)}$ is a likelihood of a (image, digit) pair for our model. Answering this query corresponds to computing the probability that the model generates the corresponding pair.  

Using standard variational inference arguments (Appendix A), we can provide a lower bound to the log probability of the pair, i.e.:

\[ \log p(X,d) \ge \log p(d) + E_{z}[\log p(X|z)] + KL(p(z|d) || q(z|X))\]

where:
\begin{itemize}
    \item $E_{z}[\log p(X|z)]$ is the likelihood of independent Gaussians (maps to an MSE, standard in VAE);
    \item $KL(p(z|d) || q(z|X))$ is the KL between the prior $p(z|d)$ and the posterior $q(z|X)$.
\end{itemize}

\section{Likelihood Lowerbound}
\label{sec:app:lowerbound}

We want to compute a lower bound to the log-likelihood of a training pair $p(X,d)$:

\begin{align*}
    \log p(X,d) & = \log \int p(X|z)p(z|d)*p(d)dz \\
    &=\log p(d) + \log\int p(X|z)p(z|d)dz
\end{align*}

Let's multiply and divide for the approximate posterior $q(z|X)$ (i.e., the encoder).

\begin{align*}
    \log p(X,d) & = \log p(d) + \log\int \frac{q(z|X)}{q(z|X)}p(X|z)p(z|d)dz
\end{align*}

For Jensen's inequality:

\begin{align*}
    \log p(X,d) &  -  \log p(d)  \\ & \ge   \int q(z|X)\log\frac{p(X|z)p(z|d)}{q(z|X)} dz \\
    & \ge  \int q(z|X)\log p(X|z)dz + \int q(z|X) \frac{p(z|d)}{q(z|X)} dz \\
    & \ge E_{z}[\log p(X|z)] - KL(p(z|d) || q(z|X))
\end{align*}

Therefore:

\[ \log p(X,d) \ge \log p(d) + E_{z}[\log p(X|z)] + KL(p(z|d) || q(z|X))\]

\section{Experiments}
\label{sec:app:experiments}
We provide an example for the declarative queries over \texttt{multi\_add/3}. 
As described, we randomly generate two $4$-digit numbers represented by a list of digit images.
These numbers have to add up to the last element of the query.
We randomly substitute $4$ digits with variables to re-generate them.
As the generated MNIST digits are hard to distinguish in this small format, we use boxed digits to represent them.

The original query 
\begin{equation}
    \begin{split}
        \texttt{multi\_add(}&\texttt{[} \boxed{1}, \boxed{4}, \boxed{3}, \boxed{8}\texttt{],}\\
        &\texttt{[}\boxed{3}, \boxed{5}, \boxed{2}, \boxed{5}\texttt{], 4963.}
    \end{split}
\end{equation}
is masked to 
\begin{equation}
    \begin{split}
        \texttt{multi\_add(}&\texttt{[} \boxed{?}, \boxed{4}, \boxed{?}, \boxed{?}\texttt{],}\\
        &\texttt{[}\boxed{?}, \boxed{5}, \boxed{2}, \boxed{5}\texttt{], 4963.}
    \end{split}
\end{equation}

And prompted to our model. 
For each answer, which is a list of generated images, we compute the labels of the closest image in the image space. 
If those labels satisfy the underlying addition, we denote a query as successfully answered.

\section{Implementation Details}
The implementation provided within the main body is great for educational purposes but is tedious to maintain once we want to change parameters.
We use a slightly different formulation for the experiments that grounds out to the same implementation but requires more Prolog knowledge.

As described, we change the probability of \texttt{decode} between training and inference. 
The full training program used is shown in \cref{app:fig:training_implementation}. 
Here, we treat all prototypes at the same time using a \texttt{map\_list}.
As in this formulation \texttt{Image} is only bound once, we have to make the list of images explicit to generate images for all possible prototypes.

\begin{figure*}[t]
    \centering

\begin{subfigure}{\textwidth}
    
    \begin{minted}[linenos, xleftmargin=20pt, highlightlines={36-37}, highlightcolor=red!15,]{prolog}
number([],Result,Result).
number([H|T],Acc,Result) :- digit(H,Nr), Acc2 is Nr+10*Acc,number(T,Acc2,Result).
number(X,Y) :- number(X,0,Y).

multi_addition(X,Y,Z) :- number(X,X2),number(Y,Y2), Z is X2+Y2.
addition(Img1,Img2,Sum) :- digit(Img1,D1), digit(Img2,D2), Sum is D2+D1.

prototype(X, tensor(prototype(X))) :- between(0,9,X). 

P0::digit(I0,0) ; P1::digit(I1,1); P2::digit(I2,2); P3::digit(I3,3); 
    P4::digit(I4,4); P5::digit(I5,5); P6::digit(I6,6); P7::digit(I7,7); 
    P8::digit(I8,8); P9::digit(I9,9):- 
        all_prob([I0,I1,I2,I3,I4,I5,I6,I7,I8,I9],[0,1,2,3,4,5,6,7,8,9],
        [P0, P1, P2, P3, P4, P5, P6, P7, P8, P9]).

maplist(_, [], []).
maplist(P, [H1|T1], [H2|T2]) :-
    call(P, H1, H2),
    maplist(P, T1, T2).

map_encode_decode([], [], []).
map_encode_decode([Image|Images], [Prot|Prototypes], [P|Probs]) :- 
    encode_decode(Image, Prot, P),  map_encode_decode(Images, Prototypes, Probs).

all_prob(Images,Classes, Dists) :- maplist(prototype,Classes,Prototypes), 
    map_encode_decode(Images, Prototypes, Dists).

encode_decode(Image, Prototype, P) :- encode(Image, Prototype, P1), 
    decode(Prototype, Image, P2), mul(P1, P2, P). 

encode(Image, Prot, P) :- ground(Image), encoder(Image,Latent), 
    lat_similar(Prot, Latent, P).
encode(Image, Prot, P) :- var(Image), sample(Prot, Sample), 
    decoder(Sample, Image), lat_similar(Prot, Sample, P).

decode(Prot, Image, P) :- ground(Prot), sample(Prot, Latent), 
    decoder(Latent, Image2), im_similar(Image, Image2, P).
decode(Prot, Image, 1.0) :- var(Prot), encoder(Image, Prot).

nn(encoder, [Image], Latent) :: encoder(Image, Latent).
nn(decoder, [Latent], Image) :: decoder(Latent, Image).

im_similar(X,X, 1.0).
im_similar(Image1, Image2, P) :- Image1 \= Image2, mse(Image1, Image2, P).

lat_similar(X,X, 1.0).
lat_similar(Lat1, Lat2, P) :- Lat1 \= Lat2, likelihood(Lat1, Lat2, P).

encode(Image, Latent, 1.0) :- var(Image), sample(Latent, Sample), 
    decoder(Sample, Image), lat_similar(Latent, Sample, P).
    \end{minted}
\caption{Full training program used for experiments.}
\end{subfigure}

\begin{subfigure}{\textwidth}
    \begin{minted}[linenos, xleftmargin=20pt, highlightlines={1}, highlightcolor=green!15,]{prolog}
decode(Prot, Image, 1.0) :- ground(Prot).
    \end{minted}
    \caption{Substitutions for the program used during inference, not using \texttt{im\_similar} anymore. 
    The brief decode is feasible, as the image is either given or already generated during encode.}
\end{subfigure}

    \caption{Full training program used for experiments. 
    While during training the red snippet is used, we use the green snippet during inference.}
    \label{app:fig:training_implementation}
\end{figure*}

\end{document}